\documentclass[letterpaper, 10 pt,conference]{ieeeconf}
\IEEEoverridecommandlockouts
\let\labelindent\relax

\usepackage{caption}
\usepackage{tikz, tikz-3dplot}

\graphicspath{{figures/}} % declare the path where graphic files are
\DeclareGraphicsExtensions{.pdf,.jpg,.png} % grafic files extensions
\usepackage{amsmath} % math package for split
\usepackage{amsfonts} % math package for fonts
\usepackage{tikz} % figure package
\usetikzlibrary{shapes,arrows,fit,calc,positioning,automata}
\usepackage{dblfloatfix} % figure package for positioning figures on the bottom
\usepackage{color} % colors package
\usepackage{multirow} % multiple rows in table
\usepackage{hhline} % separator in table
\usepackage[bookmarks=false]{hyperref} % hyperlink package
\usepackage{enumerate} % enumerate package
\usepackage{epstopdf} % package to include eps. files
\usepackage{comment}
\usepackage{svg}
\usepackage{pdfpages}
\usepackage{algorithm}
\usepackage[export]{adjustbox}
\usepackage{graphicx}
\usepackage{setspace}
\usepackage{graphicx}
\usepackage{caption}
\usepackage{subcaption}
\usepackage{amsmath}
\usepackage{amssymb}
\usepackage{algorithm}
\usepackage{algorithmic}
\usepackage{xcolor}
\usepackage{booktabs}
\usepackage{multirow}
\usepackage[table]{colortbl}% http://ctan.org/pkg/xcolor
\usepackage{diagbox}
\usepackage{makecell}
\usepackage{enumitem}
\usepackage{tikz}
\usepackage[most]{tcolorbox}
\usepackage[nolist]{acronym}
\usepackage{flushend}

\begin{acronym}[MPC]
\acrodef{ACF}{Aggregated Channel Features}
\acro{AIP}{Anterior Parietal Lobe}
\acro{ANN}{Artificial Neural Networks}
\acro{BO}{Bayesian Optimization}
\acro{CAD}{Computer-Aided Design}
\acro{CIP}{Caudal Intraparietal Sulcus}
\acro{CNN}{Convolutional Neural Network}
\acro{DNN}{Deep Neural Networks}
\acro{FOV}{Field of View}
\acro{DCNN}{Deep Convolutional Neural Network}
\acro{GMM}{Gaussian Mixture Model}
\acrodefplural{GMM}{Gaussian Mixture Models}
\acro{GP}{Gaussian Proccess}
\acrodefplural{GP}{Gaussian Proccesses}
\acro{GPU}{Graphical Proccessing Unit}
\acrodefplural{GPU}{Graphical Proccessing Units}
\acro{GHT}{Generalized Hough Transform}
\acro{HOG}{Histogram of Gradients}
\acro{HRI}{Human Robot Interaction}
\acro{HVS}{Human Visual System}
\acro{IOR}{Inhibition of Return}
\acro{KL}{Kullback-Leibler}
\acro{JPDA}{Joint Probabilistic Data Association}
\acro{LGN}{Lateral Geniculate Nucleus}
\acrodefplural{LSTM}[LSTM]{Long Short-Term Memory Units}
\acro{MAB}{Multi-Armed Bandit}
\acro{MAP}{Maximum a Posteriori}
\acro{MCTS}{Monte Carlo Tree Search}
\acro{MOT}{Multiple Object Tracking}
\acro{NBV}{Next-Best-View}
\acro{POMDP}{Partially Observable Markov Decision Process}
\acrodefplural{POMDP}{Partially Observable Markov Decision Processes}
\acrodefplural{GMM}[GMM]{Gaussian Mixture Models}
\acro{RANSAC}{Random Sample Consensus}
\acro{RF}{Receptive Field}
\acrodefplural{RF}{Receptive Fields}
\acro{RNN}{Recurrent Neural Network}
\acro{RPN}{Region Proposal Network}
\acro{SES}{Sensory Ego-Sphere}
\acro{SIFT}{Scale Invariant Feature Transform}
\acrodef{SVM}{Support Vector Machine}
\acrodefplural{SVM}{Support Vector Machines}
\acro{UCB}{Upper Confidence Bound}
\acrodefplural{UCB}{Upper Confidence Bounds}
\acro{UT}{Unscented Transform}
\acro{WTA}{Winner Take All}
\acro{RoI}{Region of Interest}
\acro{FPS}{frames per second}
\acro{MOTP}{Multiple Object Tracking Precision}
\acro{SLAM}{simultaneous localization and mapping}
\acro{RRT}{Rapid random tree}
\acroplural{RRT}[RRTs]{rapid random trees}
\acro{UAV}{Unmanned aerial vehicle}
\acroplural{UAV}[UAVs]{Unmanned aerial vehicles}
\acro{IMU}{Inertial Motion Unit}
\acroplural{IMU}[IMUs]{Inertial Motion Units}
\acro{VIO}{Visual Inertial Odometry}
\end{acronym}

\begin{document}
\title{On the Advantages of Multiple Stereo Vision Camera Designs for Autonomous Drone Navigation}
\author{
Rui Pimentel de Figueiredo, Jakob Grimm Hansen, Jonas Le Fevre, Martim Brand\~ao, Erdal Kayacan
\thanks{R. Figueiredo, J. Hansen, J. Fevre, E. Kayacan are with Artificial Intelligence in Robotics Laboratory (Air Lab), the Department of Electrical and Computer Engineering, Aarhus University, 8000 Aarhus C, Denmark
        {\tt\small \{rui,jakob,jonas.le.fevre,erdal\} at ece.au.dk}}
\thanks{M. Brandao is with King's College London (KCL), London, UK
        {\tt\small \{martim.brandao\} at kcl.ac.uk}}
}

% use a multiple column layout for up to three different affiliations

\maketitle

\begin{abstract}

In this work we showcase the design and assessment of the performance of a multi-camera UAV, when coupled with state-of-the-art planning and mapping algorithms for autonomous navigation. The system leverages state-of-the-art receding horizon exploration techniques for \ac{NBV} planning with 3{D} and semantic information, provided by a reconfigurable multi stereo camera system. We employ our approaches in an autonomous drone-based inspection task and evaluate them in an autonomous exploration and mapping scenario. We discuss the advantages and limitations of using multi stereo camera flying systems, and the trade-off between number of cameras and mapping performance. 

%\textcolor{red}{Must say that semantics are expensive and clever use of processing is a must.}

\end{abstract}

%\IEEEpeerreviewmaketitle

\section{Introduction}
\label{sec:intro}

\acp{UAV} deployed in everyday environments are facing increasingly complex scenarios and tasks. The problem of selecting which regions of the surrounding environment to attend to during visual exploration, search, and mapping tasks is computationally and energetically demanding. Therefore, \acp{UAV} should be endowed with efficient active perception mechanisms that allow them to attend to objects of interest while avoiding processing irrelevant sensory information. Furthermore, the design of systems with perceptual redundancy are of utmost importance in order to ensure safety and robustness to failures, since sensor arrays can significantly improve perceptual coverage, task-execution speed, and overall state estimation accuracy. Hence, the designer of the robotic system should carefully select an appropriate number and type of sensors, taking into account task performance as well as on-board resource-constraints.% (i.e. computational and energetic).
\begin{figure}[ht!]
  \centering
  \begin{subfigure}[b]{0.35\textwidth}
  	\includegraphics[width=1.0\textwidth]{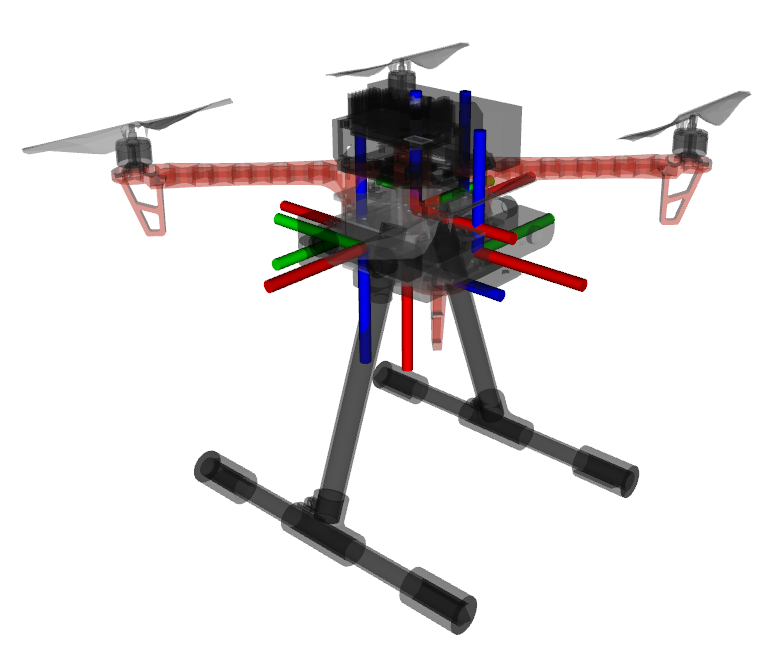}
	\caption{Multi Stereo Camera UAV CAD design}
  \end{subfigure}
  \begin{subfigure}[b]{0.18\textwidth}
	\includegraphics[width=1.0\textwidth]{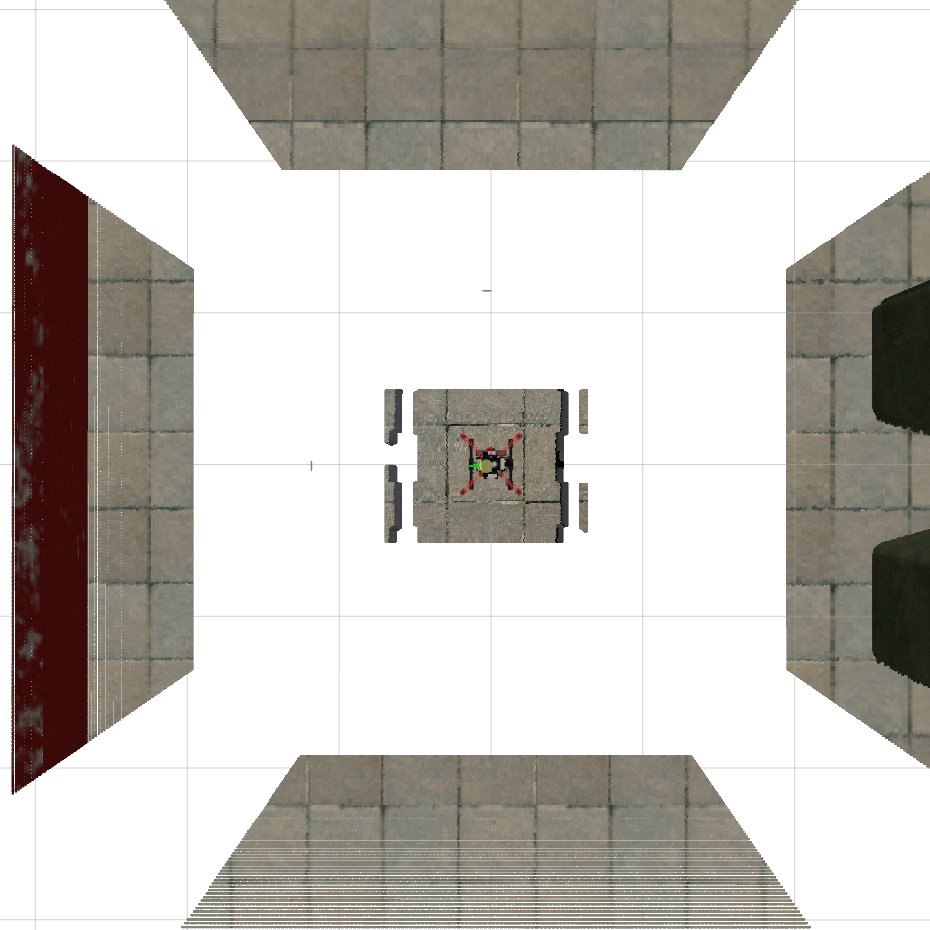}
	\caption{RGBD}
  \end{subfigure}
  \begin{subfigure}[b]{0.18\textwidth}
	\includegraphics[width=1.0\textwidth]{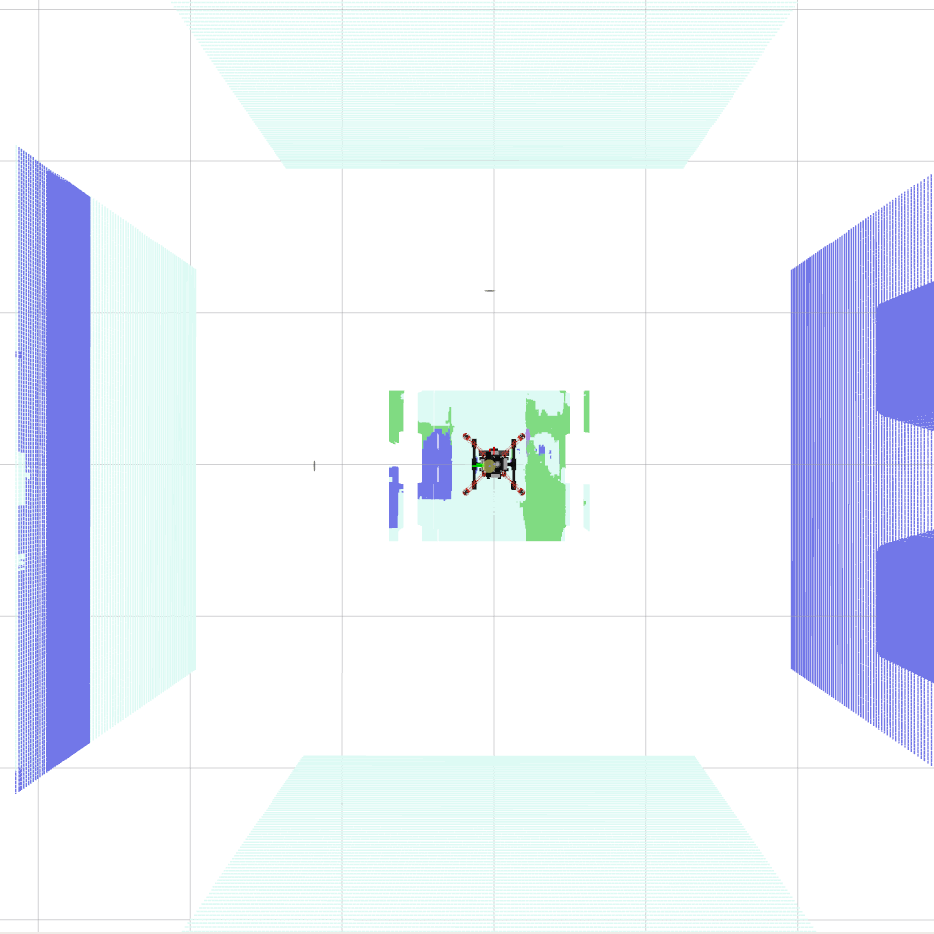}
	\caption{Semantics}
  \end{subfigure}
  \caption{The proposed multi stereo camera UAV platform for autonomous drone navigation applications. The design allows to easily attaching a total of 5 stereo vision cameras around the UAV frame (i.e. front, left, right, back, bottom).} 
  \label{fig:multi_stereo_camera_system} 
\end{figure}
High perceptual coverage for safe navigation and mapping of real world scenarios can be achieved using a flying drone vehicle equipped with vision and IMU systems. Using multiple cameras and IMUs offers a robust solution. However, it comes at the cost of increased payload, and additional computational power and processing-time requirements. In this work we asses the viability of multi-stereo-camera \ac{UAV} (see Fig. \ref{fig:multi_stereo_camera_system}) for autonomous inspection tasks, combining state-of-the-art \ac{SLAM} techniques, with cost-efficient \ac{NBV} exploration algorithms \cite{bircher2018receding}, to geometrically reconstruct and label all objects in man-made environments. %
Our navigation system is targeted at multi-camera \acp{UAV}, includes probabilistic semantic-metric mapping representations, and uses a RRT planning algorithm that leverages both semantic and metric information for autonomous visual data collection. Our target application is the inspection of man-made structures, requiring minimal human intervention.
Throughout the rest of this article we overview the proposed system design and perform an evaluation of the advantages and disadvantages of using multi-camera systems using \acp{UAV}, from a computational and mapping performance perspective. This work assesses the former problem trade-offs on an UAV-based exploration and mapping scenario.
%To our knowledge this study is of the utmost importance for resource-constrained autonomous robotic applications, and 
%The results section \ref{sec:results} evaluates the proposed strategies for autonomous mapping and navigation, in a realistic simulated shipyard environment. Finally, we discuss our main results, limitations and future work in the final conclusions section.

\begin{figure*}[ht!]
  \centering
  \includegraphics[width=1.0\textwidth]{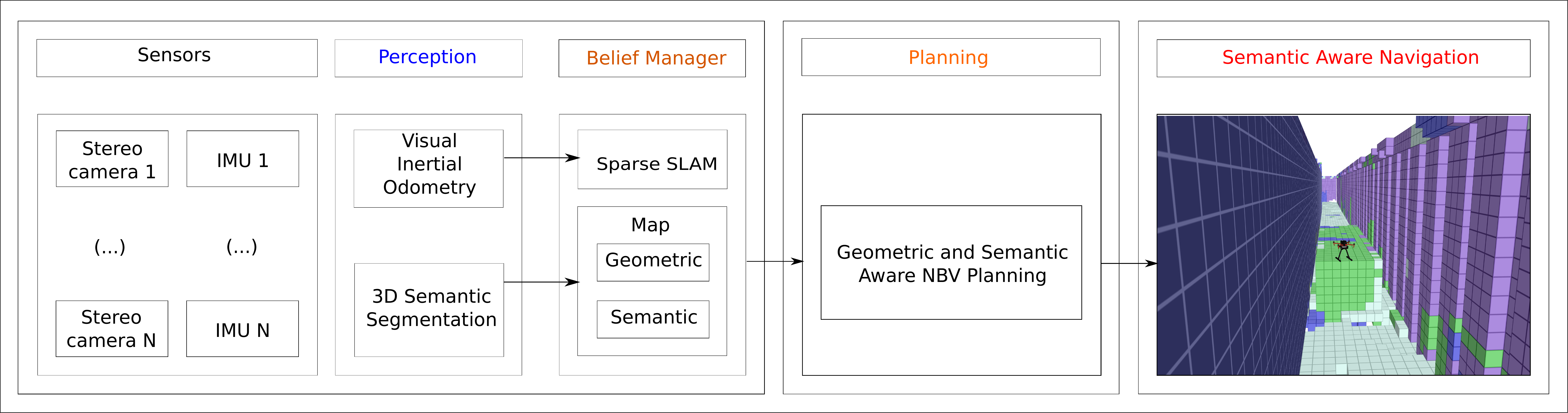}
  \caption{Overview of the proposed autonomous navigation system for localization and semantic-metric mapping of man-made environments.}
  \label{fig:shipyard_inspection_pipeline}
\end{figure*}

\section{Methodology}
\label{sec:methodology}
In the rest of this section we describe the proposed multi-stereo-camera system and methodologies for active exploration and semantic-metric mapping of man-made infrastructures.

\subsection{System Overview}

The proposed system for autonomous navigation tasks
%is depicted in Fig. \ref{fig:shipyard_inspection_pipeline}, and 
consists of a \ac{UAV} specifically designed for mapping tasks, that comprises multiple cameras, \acp{IMU}, and an altimeter. 
Our navigation system relies on an off-the-shelf \ac{SLAM} system with loop closing and relocalization capabilities \cite{murORB2}, which is fed with RGB-D data provided by user-selected cameras %and \ac{VIO} estimates, obtained from the cameras 
and \acp{IMU} measurements. These are fused using an extended Kalman filter (EKF) for improved robustness on self-motion tracking performance.% \cite{rovio2015}. 
In the proposed hardware design we attempt to minimize weight to achieve better flight performance and flight duration until battery depletion, while at the same time minimizing the size of the parts to avoid vibrations, and ensure camera sensors can be placed in orthogonal directions to maximize visual coverage. Our model adds camera sensors to the quad base frame (DJI F450), together with a battery mount for easy battery replacement. We use Jetson Xavier NX as our on-board computer and the Pixhawk 4 as the low-level flight controller. Furthermore, the system comprises a set of stereo-camera sensors (Zed 2 and Mini cameras) suitable for visuo-inertial based navigation which are rigidly attached to the UAV body base frame, and whose poses are assumed deterministically known with respect to the base frame, from the kinematics model. 

%\begin{figure}[h]
%  \centering
%  \includegraphics[width=0.5\textwidth]{images/uav_design_small.jpg}
%  \caption{Drone Hardware System Design}
%  \label{fig:drone_hw_implementation}
%\end{figure}
\begin{comment}
\begin{table}[]
\centering
\small
\begin{tabular}{lll}
Components & Quantity & Weight (g/pcs) \\ \hline
Holybro Pixhawk 4 (Pix4+PM07) & 1 & 15.8 + 36 \\
DJI Framewheel 450 & 1 & 282 \\
DJI propeller (24x12.7 cm) & 4 & 13 \\
DJI 2312E 800KV Motor & 4 & 56 \\
DJI 430 Lite ESC & 4 & 11.6 \\
Jetson Xavier (NX+NGX004) & 1 & 85 + 33 \\
ZED2 & 1 & 166 \\
ZED mini & 1 to 4 & 62.9 \\
LiPo Battery (4s 5000mAh) & 1 & 450 \\
3D Printed Parts & 1 & 200 \\ \hline
UAV Total Weight &  & 1675.1
\end{tabular}  
\caption{Drone Hardware Components}
\label{fig:drone_hw_implementation}
\end{table}
\end{comment}
\subsubsection{Multi-camera Navigation System}

We rely on a probabilistic observation model that combines metric and semantic visual cues, which are efficiently fused in a volumetric octogrid structure \cite{hornung2013octomap}, and a \ac{NBV} planner that leverages both geometric and semantic information for task-dependent exploration.
%We consider an octomap representation, defined as a 3D uniform voxel grid structure that encloses the workspace around the robot, represented by $m=\{m_i\}$, where each voxel $m_i=\{m^o_i, m^s_i\}$ with $m^o_i\in \{0,1\}$ being a binary random variable representing its occupancy, and $m^s_i\in \{1,...,K_c\}$ a semantic variable representing its object class. 
We use an octomap representation and recursive Bayesian volumetric mapping%~\cite{thrun2005probabilistic}
to sequentially estimate the posterior probability distribution over the map, given sensor measurements  and sensor poses obtained through the robot kinematics model and an off-the-shelf SLAM module. %Filtering updates are recursively computed in log-odds space to ensure numerical stability and efficiency. %~\cite{moravec1985high} 
Our method for semantic segmentation relies on a \ac{DCNN} encoder-decoder segmentation network, that receives RGB or grayscale images as input, and outputs a probability distribution over the known object categories for each pixel $(u,v)$. We use BiseNet \cite{bisenet2018} because it is compact, fast, robust and easy to use, being suitable for remote sensing applications running on embedded systems (e.g. UAVs) with low computational specifications. For each pixel $(u,v)$, the network outputs a probability distribution $p^c(u,v)\in\mathcal{P}^{K_c}$ over the set of known classes $\mathcal{C}$, where $K_c$ represents the number of known classes. For training the network we use a combination of real and simulated (AirSim) annotated datasets, and the categorical Cross-Entropy loss function.
%\begin{align}
%    CE = -\text{log}\left(\frac{e^{s_p}}{\sum_{j}^C e^{s_j} }\right)
%\end{align}
%where $s_j$ is the \ac{CNN} output score for the class $j\in \mathcal{C}$, and $s_p$ the positive class.
At run-time, the semantic probability distribution over all classes and image pixels is merged with the corresponding depth image to obtain a semantically labeled point cloud, using a known extrinsic calibration parametric model. %(see Figure \ref{fig:semantic_pcl}).

\section{Results}
\label{sec:results}

\begin{figure}[t]
    \centering
    \begin{subfigure}[b]{0.21\textwidth}
    	\includegraphics[width=1.0\textwidth]{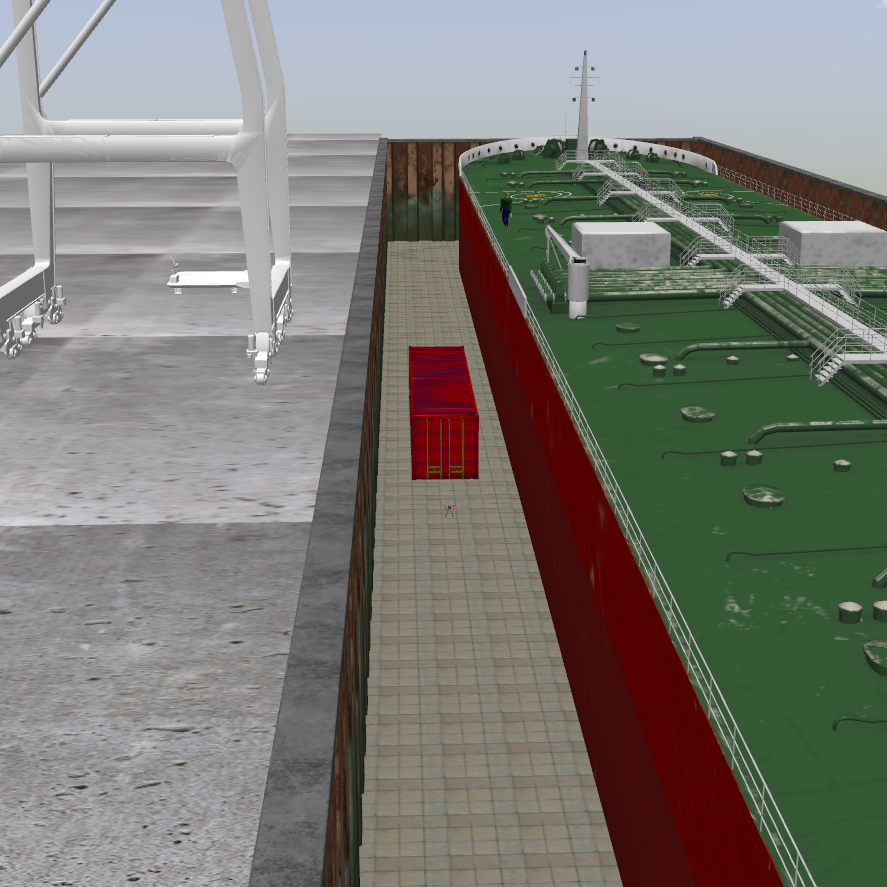}
    	\caption{Dry-dock gazebo environment.}
	    \label{fig:gazebo_dry_dock}
    \end{subfigure}    
    \begin{subfigure}[b]{0.23\textwidth}
    	\includegraphics[width=1.0\textwidth]{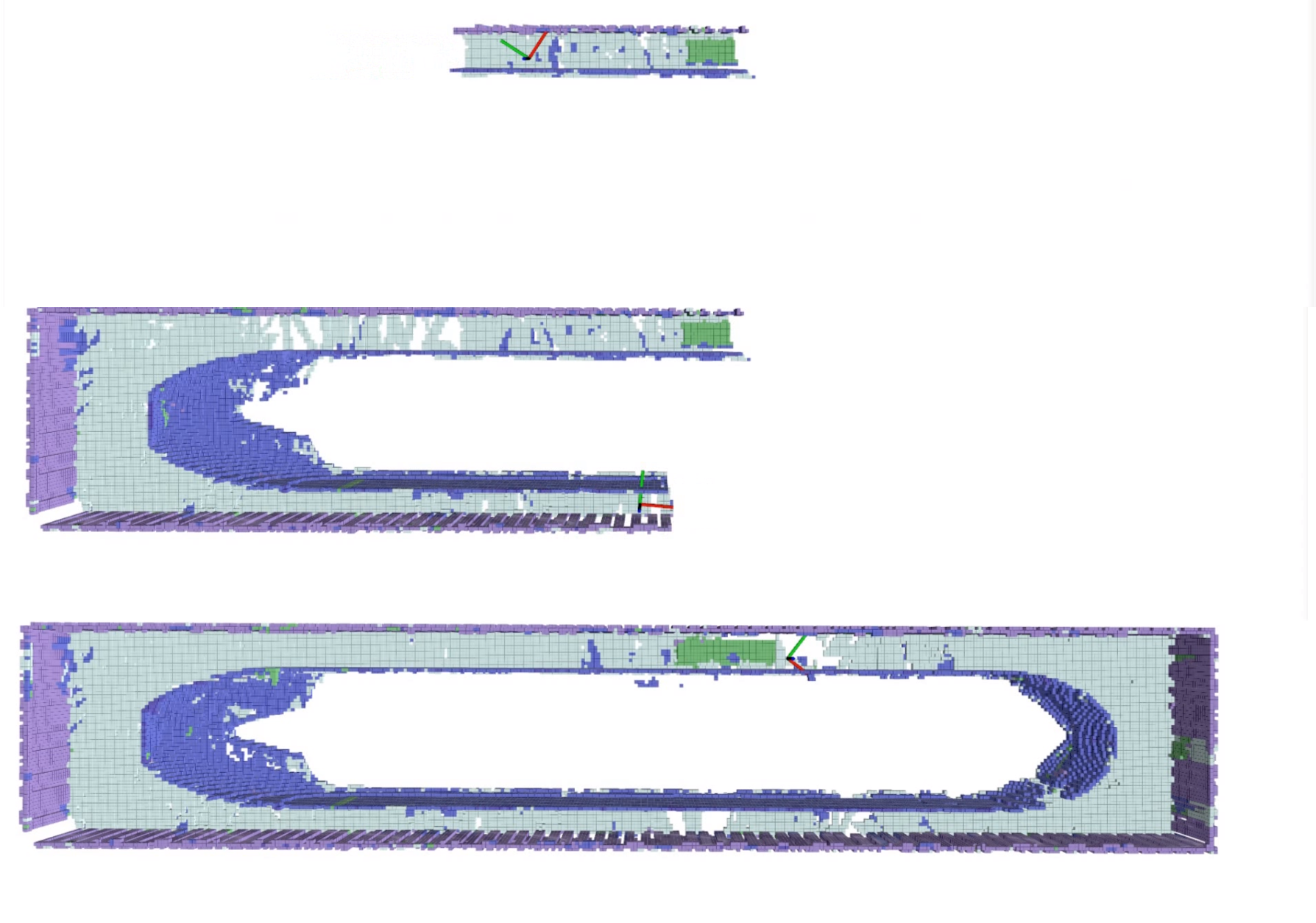}
    	\caption{Octomap is colored according to most likely semantics.}
    \end{subfigure}
    \begin{subfigure}[b]{0.45\textwidth}
	\includegraphics[width=1.0\textwidth]{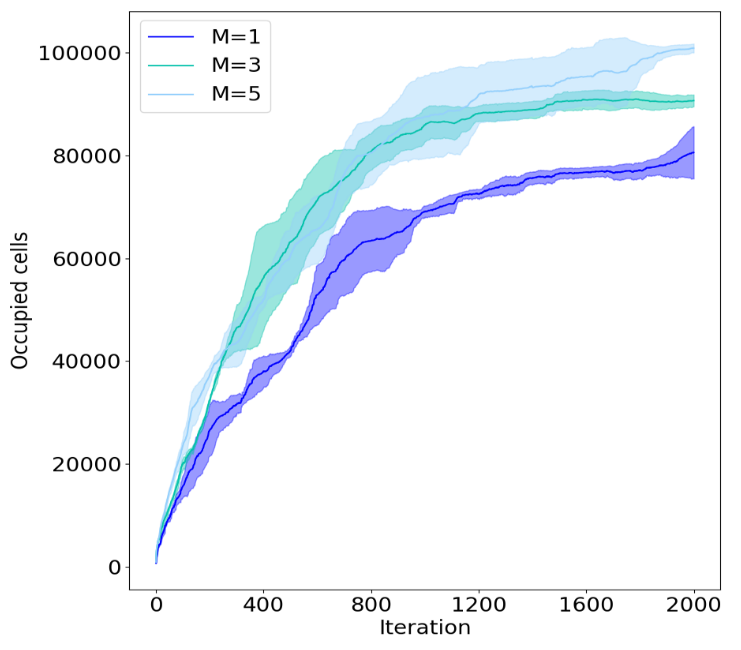}
        \caption{Evolution of the average number of occupied cells.}
    \end{subfigure}	
    \caption{Mapping performance temporal evolution of our multi stereo camera system in a realistic simulation environment.} %Each iteration includes planning, act, and sensing acquisition and fusion in the volumetric-semantic grid.}
    \label{fig:overall_simulation_results}
\end{figure}

\begin{figure}[t]
  \centering
  \begin{subfigure}[b]{0.48\textwidth}
    \includegraphics[width=1.0\textwidth]{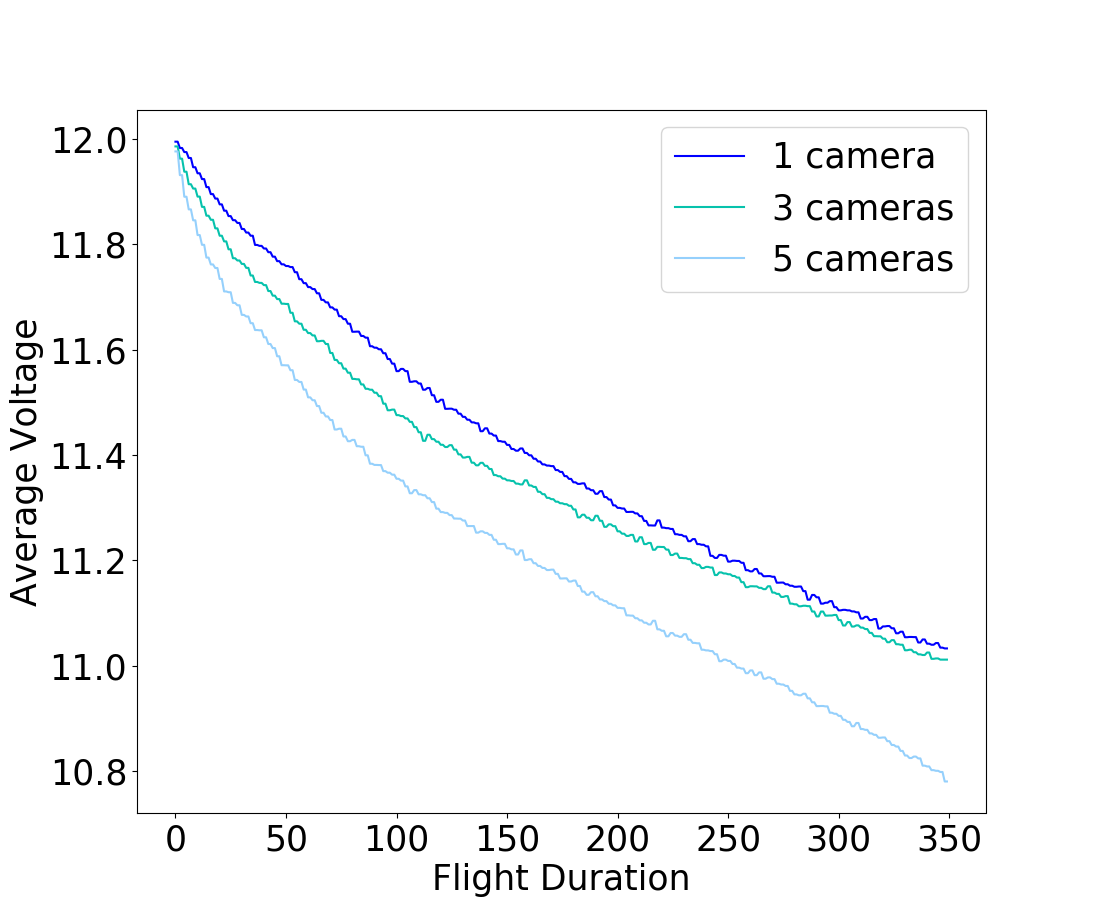}
  \end{subfigure}
  %\begin{subfigure}[b]{0.48\textwidth}
  %  \includegraphics[width=0.48\textwidth]{images/indoor_hovering.png}
  %\end{subfigure}
  \caption{Battery consumption profile.}% (over planning iterations).}   
  \label{fig:battery_performance} 
\end{figure}

%\begin{figure}[t]
%  \centering
%\begin{subfigure}[b]{0.48\textwidth}
%	\includegraphics[width=0.48\textwidth]{images/occupancy.png}
%	\caption{}
%\end{subfigure}
%\caption{Temporal evolution of the overall mapping occupancy for our system, running in a realistic simulation environment, with different number of cameras. %Each iteration includes planning, act, and sensing acquisition and fusion in the volumetric grid.
%}
%\label{fig:overall_simulation_results}
%\end{figure}

\subsection{Multi-camera Navigation System}
In order to be able to quantitatively and qualitatively measure the performance of the proposed mapping and planning approaches, a realistic shipyard environment (see Fig. \ref{fig:gazebo_dry_dock}) was created using the Gazebo simulator \cite{gazeboSimulationEnvironment}. The environment consists of a dry-dock. An intelligent active mapping algorithm should maximize task-related rewards, in this case information gathering, by focusing on rewarding viewing directions. In each experiment we let the observer collect $T=2000$ observations (i.e.sense, plan and act iterations). Each experiment was repeated 10 times to average out variability in different simulations, due to the randomized nature of our algorithm, and non-systematic errors influenced by multiple simulation factors.

We first analyzed the influence of different camera setups in the trade-off between reconstruction accuracy, planning, and run-time performance. For the number of cameras, we considered $M\in\{1;3;5\}$. Fig.~\ref{fig:overall_simulation_results} demonstrates the advantages of utilizing multiple cameras placed around the UAV. For this particular scenario, on average, the use of multiple cameras not only improves occupancy but also the time-to-full-coverage. However, the cameras placed on the back and bottom provide lower long-term information when compared to the front and lateral cameras, since most of the surrounding environment is covered by the latter while the system is moving.

\subsection{Multi-Camera Drone Hardware Design}
In order to select the most suitable camera configuration for our autonomous flying system, we measured battery power consumption and time-to-full-coverage across multiple designs (i.e. different number of sensors) while hovering the real UAV with a different number of cameras (for 5 different runs). As can be seen in Fig. \ref{fig:battery_performance}, the power consumption (proportional to lithium battery voltage) increases with the number of cameras since they increase both the weight and processing requirements of the system. Hence, although higher visibility and faster coverage can be achieved with more cameras ($M=5$), when considering power constraints and flight duration (Table. \ref{tab:flight_time}), $M=3$ is a more appropriate design choice for this use-case.
\begin{table}[h]
\small
\centering
\begin{tabular}{|c|c|c|c|}
\hline
Cameras               & $M=1$     & $M=3$     & $M=5$     \\ \hline
Flight Time & $8.79\pm0.86$ & $8.17\pm0.41$ & $6.00\pm0.99$ \\ \hline
\end{tabular}
  \caption{Hovering time until battery depletion (minutes).}   
  \label{tab:flight_time} 
\end{table}
\section{Conclusions}
In this work we have proposed and assessed multi-stereo-vision camera setups for autonomous navigation of \acp{UAV} that incorporates probabilistic semantic-metric mapping representations, for semantically-aware \acp{NBV} planning. We assessed the proposed designs and methodology on a realistic simulation environment (Gazebo), and evaluated the trade-offs of using multi-camera navigation systems in UAV-based inspection tasks. Our final design choice considered power, computation requirements and flight-time duration, in a real experimental setup. In the future we intend to improve the proposed multi-camera approach with the ability to schedule sensor acquisition such as to decrease computational load and power consumption.

\section*{Acknowledgment}
The authors would like to acknowledge the financial contribution from Smart Industry Program (European Regional Development Fund and Region Midtjylland, grant no.: RFM-17-0020). The authors would further like to thank Upteko Aps for bringing use-case challenges.

\bibliographystyle{IEEEtran}

%=================== for overleaf
%
%\bibliography{References}

%=================== for arxiv

\end{document}